\pdfoutput=1

\documentclass[11pt]{article}

\usepackage[]{acl}

\usepackage{times}
\usepackage{latexsym}

\usepackage[T1]{fontenc}

\usepackage[utf8]{inputenc}

\usepackage{microtype}

\usepackage{inconsolata}
\usepackage{graphicx}
\usepackage{booktabs}
\usepackage{multirow}
\usepackage{xcolor} 
\usepackage{listings}
\usepackage{hyperref}
\usepackage{multicol} 
\usepackage{lipsum} 
\usepackage{CJKutf8}

\title{Integrate the Essence and Eliminate the Dross: Fine-Grained Self-Consistency for Free-Form Language Generation}

\begin{document}
\begin{CJK}{UTF8}{gbsn}
\author {
    \textbf{Xinglin Wang}\textsuperscript{\rm 1}\footnotemark[1], \hspace{0cm}
    \textbf{Yiwei Li}\textsuperscript{\rm 1}\footnotemark[1], \hspace{0cm}
    \textbf{Shaoxiong Feng}\textsuperscript{\rm 2}, \hspace{0cm} 
    \textbf{Peiwen Yuan}\textsuperscript{\rm 1}, \hspace{0cm} 
    \textbf{Boyuan Pan}\textsuperscript{\rm 2}, \hspace{0cm} \\
    \textbf{Heda Wang}\textsuperscript{\rm 2}\textbf{,} \hspace{0cm} 
    \textbf{Yao Hu}\textsuperscript{\rm 2}\textbf{,} \hspace{0cm} 
    \textbf{Kan Li}\textsuperscript{\rm 1}\footnotemark[2] \\
    \textsuperscript{\rm 1} School of Computer Science, Beijing Institute of Technology \\
    \textsuperscript{\rm 2} Xiaohongshu Inc \\
    \texttt{\{wangxinglin,liyiwei,peiwenyuan,likan\}@bit.edu.cn} \\
    \texttt{\{shaoxiongfeng2023,whd.thu\}@gmail.com} \  \texttt{\{panboyuan,xiahou\}@xiaohongshu.com}
}

\maketitle

\renewcommand{\thefootnote}{\fnsymbol{footnote}} 
\footnotetext[1]{Equal contribution.} 
\footnotetext[2]{Corresponding author.} 

\renewcommand{\thefootnote}{\arabic{footnote}}

\begin{abstract}
Self-consistency (SC), leveraging multiple samples from LLMs, shows significant gains on various reasoning tasks but struggles with free-form generation due to the difficulty of aggregating answers.
Its variants, UCS and USC, rely on sample selection or voting mechanisms to improve output quality. These methods, however, face limitations due to their inability to fully utilize the nuanced consensus knowledge present within multiple candidate samples, often resulting in suboptimal outputs. 
We propose Fine-Grained Self-Consistency (FSC) to addresses these limitations by extracting and integrating segment-level commonalities from candidate samples, enhancing the performance of LLMs both in open-ended and reasoning tasks. 
Based on this, we present two additional strategies: candidate filtering, which enhances overall quality by identifying highly similar candidate sets, and merging, which reduces input token requirements by combining similar samples. 
The effectiveness of FSC is demonstrated through extensive experiments on various tasks, including summarization, code generation, and mathematical reasoning, using GPT-3.5-turbo and GPT-4. The results indicate significant improvements over baseline methods, showcasing the potential of FSC to optimize output quality by effectively synthesizing fine-grained consensus knowledge from multiple samples\footnote{Our code and data have been released on \url{https://github.com/WangXinglin/FSC}}.
\end{abstract}
\section{Introduction}
The remarkable success of large-scale language models (LLMs) have transformed the landscape of natural language processing, showcasing significant improvements across a wide range of tasks, from reasoning tasks \citep{COT} with distinct answers like arithmetic and commonsense reasoning to free-form generation tasks like code generation \citep{code-intro} and summarization \citep{sum-intro}.

However, LLMs may still generate suboptimal samples in challenging tasks. Efforts to improve output quality involve selecting the best response from multiple samples based on specific criteria. This includes using trained models for reranking outputs \citep{ravaut-etal-2023-unsupervised} and employing LLMs to evaluate the responses \citep{liu-etal-2023-g}. However, both approaches require additional models and overlook the knowledge present among the candidates.
\citet{SC} introduce self-consistency (SC) to improve performances without additional models, which mitigates noise from individual sampling by employing a voting mechanism across multiple samples. Unfortunately, SC is limited to tasks where the final answer can be aggregated through precise matching. How to aggregate the answers for free-form problems remains unclear.

\begin{figure*}[ht]
\begin{center}
\includegraphics[width=1.0\textwidth]{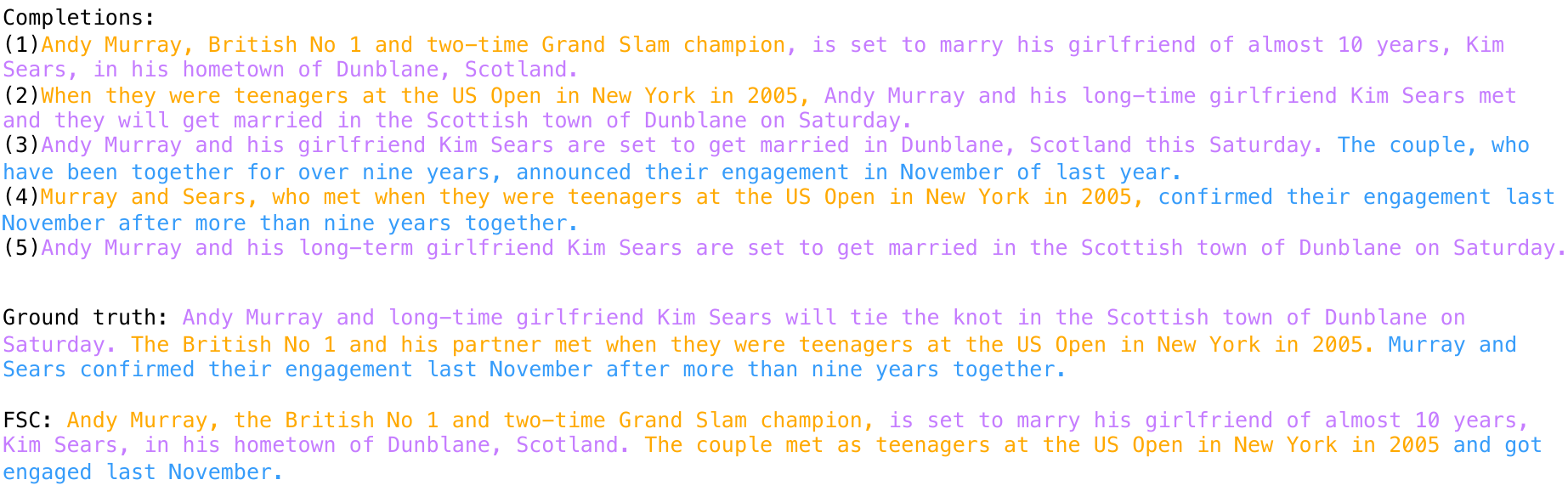}
\end{center}
\caption{Case from DailyMail with GPT-3.5-turbo. Each color represents a distinct semantic information. The input document is provided in the Appendix~\ref{ap:input} due to space constraints.}
\label{fig:intro_case}
\end{figure*}

Recently, some works seek to evolve the idea of self-consistency into open-ended generative tasks. UCS \citep{UCS} calculates the overlap of unigrams between candidates and then selects the final answer with highest value. Alternatively, USC \citep{USC} leverages the capabilities of LLMs instead of rule-based criteria to choose the most consistent one.
However, both approaches still rely on selection or voting mechanisms, which do not align well with the nature of free-form generation tasks, i.e., the final quality is determined by the entirety of the output content, rather than specific individual tokens. Therefore, sample-level selection methods only yield suboptimal outputs, primarily due to two reasons: (1) They are unable to incorporate consensus knowledge from unselected samples, which, despite their lower overall quality, may contain locally valuable information to enhance the quality of selected samples. (2) They cannot eliminate low-quality segments from selected samples to further improve overall quality.
In a word, such coarse-grained selection methods suppress or overlook the role of fine-grained consensus within the candidate samples.
Figure~\ref{fig:intro_case} and Table~\ref{tb:intro} depict scenarios that select-based SC methods struggle to address.
The former indicates that both methods perform poorly when the quality of candidates is low for code generation tasks. The latter illustrates with case the scenario where, in summarization tasks, the semantic information of the ground truth cannot be fully covered by any one of candidate samples. Thus, regardless of the selection, only suboptimal outputs can be obtained.

\begin{table}[t]
    \centering
    \small
    \begin{tabular}{c l l l}
    \toprule
        Distribution & USC & UCS  &Random \\\toprule
        5 / 0  &0.0 \% & 0.0 \% & 0.0 \% \\
        4 / 1  &13.9 \%& 13.9 \%& 20.0 \% \\
        \bottomrule
    \end{tabular}
    \caption{Accuracy on code generation benchmark HumanEval with GPT-3.5-turbo. We generate 5 samples for each problem. The term "5/0" distribution pertains to cases where all five generated codes are erroneous, while "4/1" distribution indicates that one sample is correct while the other four are incorrect.}
    \label{tb:intro}
\end{table}

To address this issue and better leverage the consensus knowledge among multiple samples for open-ended problems, we propose Fine-Grained Self-Consistency (FSC). Specifically, after generating multiple candidate samples, FSC extracts the segment-level common elements within them by taking full advantage of LLM's text comprehension and contrasting capabilities. Subsequently, it synthesizes these consensus elements to produce an optimized output, effectively integrating the essence from the candidate samples.
Based on this, we further propose two strategies: candidates filtering and merging. The former employs automated metrics to identify candidate sets that exhibit high similarity, thereby enhancing the overall quality of the candidates. The latter strategy merges samples that are highly similar, effectively reducing the quantity of input tokens required.

A wide range of tasks, including summarization, code generation, and formal mathematical reasoning, are evaluated on GPT-3.5-turbo and GPT-4 for proposed FSC. The results show that proposed method outperform baselines in a large margin. Additional experiments indicate that filter strategy can further enhance performance by selecting better candidates and merge strategy can reduce cost while maintaining the performance.

To summarize, this work includes the following contributions:
\begin{itemize}
    \item We propose Fine-Grained Self-Consistency, designed to fully leverage the consensus knowledge extracted within multiple samples for tasks involving free-form generation.
    \item Two strategies, candidates filtering and merge, are devised to improve performance and minimize costs.
    \item Extensive experiments show that our proposed method can surpasses competitive baselines.
\end{itemize}

\begin{figure*}[ht]
\begin{center}
\includegraphics[width=1.0\textwidth]{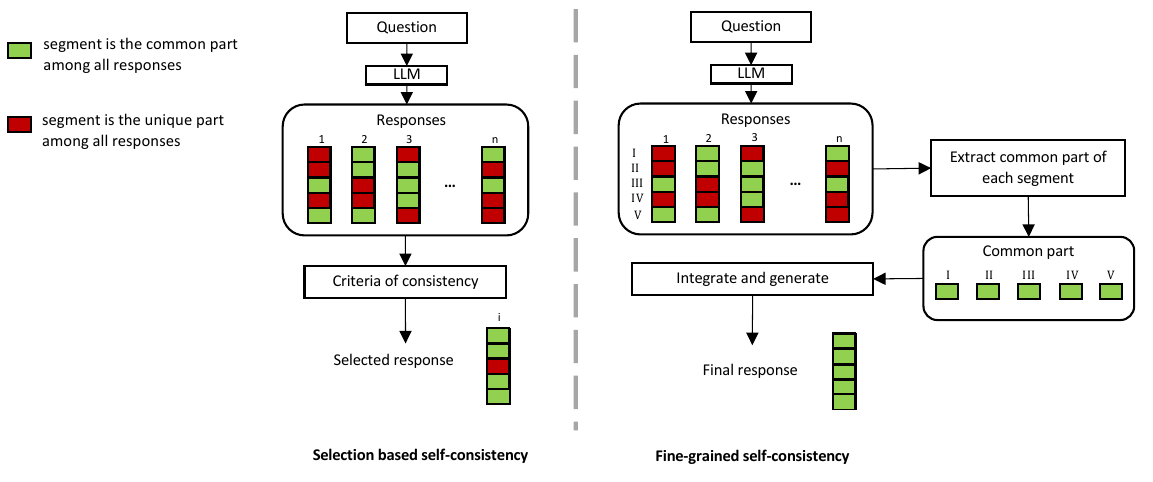}
\end{center}
\caption{Illustration of selection based self-consistency and proposed fine-grained self-consistency. }
\label{fig:method_overview}
\end{figure*}

\begin{figure*}[ht]
\begin{center}
\includegraphics[width=1.0\textwidth]{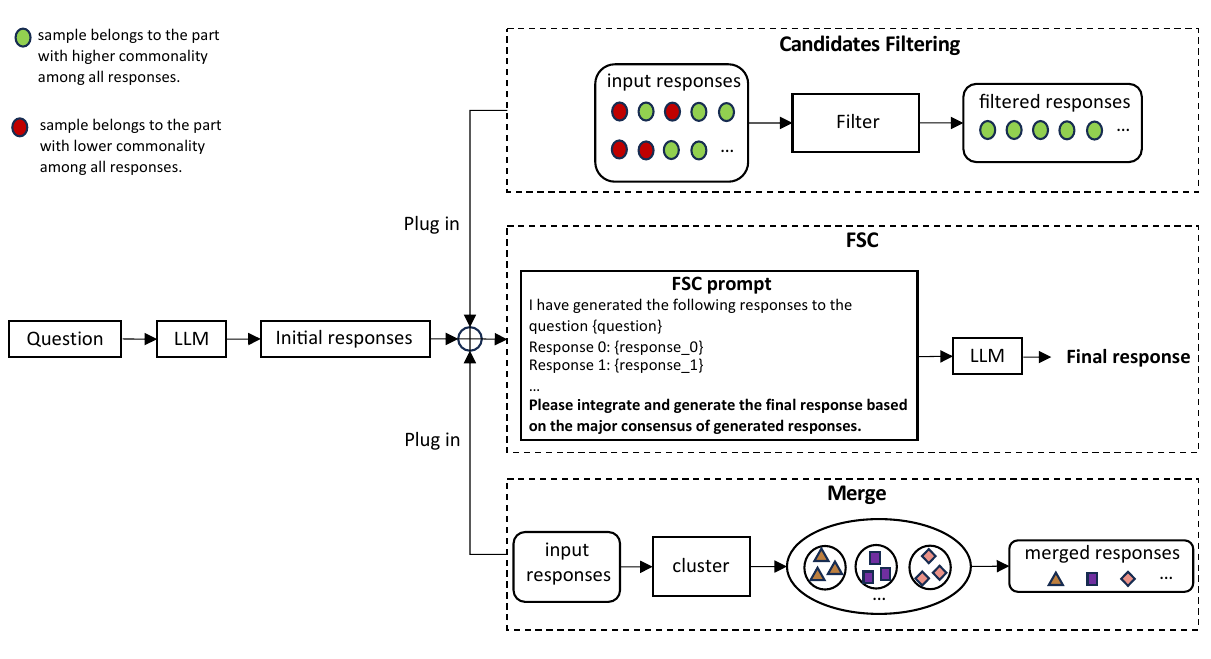}
\end{center}
\caption{Overview of the proposed fine-grained self-consistency (FSC) workflow.}
\label{fig:method_workflow}
\end{figure*}
\section{Related Work}
\paragraph{Consistency-Based Response Selection Approaches.} The literature presents a variety of consistency-based response selection approaches, typically incorporating a voting procedure to select the most frequently occurring response \citep{SC, zhou-etal-2022-prompt, portillo-wightman-etal-2023-strength, yue2023large, li2023escape}. The self-consistency approach proposed by \citet{SC} demonstrates that by generating multiple responses for the same task and selecting the reasoning path leading to the most common final answer, the performance of chain-of-thought reasoning can be improved. Candidate responses can also be derived from different prompt variants corresponding to the same problem \citep{zhou-etal-2022-prompt, portillo-wightman-etal-2023-strength, yue2023large}. For open-ended generation tasks, \citet{UCS} propose the n-gram consistency score to measure the pairwise similarity between candidate responses. The consistency score for each response is calculated as the sum of the pairwise similarity scores. \citet{USC} propose leveraging LLMs themselves to directly select the most consistent answer among multiple candidates without an explicit definition of the pairwise similarity. In this work, we take a closer look into the coarse-grained limitations faced by selection based self-consistency methods on open-ended generation tasks, and propose Fine-Grained Self-Consistency to fully leverage the segment-level consensus knowledge extracted within multiple samples so as to gain better consistency output.

\paragraph{Response improvement with multiple candidates.}
Recent studies show that the LLM can enhance its own prediction output based on candidate responses. \citet{zheng2023progressive} demonstrate that the given a trajectory of previously generated solutions, the LLM can iteratively produce superior solutions for an optimization task. Other researches focus on aggregating multiple reasoning chains and prompting the LLM to generate a better final response, which shows performance enhancement in multi-hop question answering \citep{yoran2023answering}, mathematical reasoning \citep{yang2023large}, machine translation \citep{self-contrast} and code generation \citep{multi}. Instead of prompting the LLM to generate a superior response, our proposed FSC focuses on extracting and integrating consistency, eliminating the need for post-answer guidance. We demonstrate that FSC effectively leverages multiple responses to enhance performance across a range of tasks.
\section{Method}

The core idea behind fine-grained self-consistency is to measure and extract the commonality of each response generated by LLM at the segment level, and then fuse to generate the final response with superior commonality. Figure~\ref{fig:method_overview} illustrates the idea of select-based self-consistency and the proposed fine-grained self-consistency (FSC). Specifically, select-based self-consistency ranks the responses generated by LLM according to specific consistency criteria \citep{UCS, USC}, and the top-ranked response is selected as the output. However, these methods do not assess the consistency at a more fine-grained level within the response. On the other hand, FSC first measures the commonality of each segment of the responses generated by LLM, and extracts the corresponding common part, then fuses the common part and generates the final output, achieving a more refined consistency sample output.

We present the overall workflow of Fine-grained self-consistency (FSC) in Figure~\ref{fig:method_workflow}, Including FSC with two plugins, candidates filtering and merge.

\subsection{Fine-Grained Self-Consistency}
Considering that it is difficult to divide and measure the consistency of the segments in the responses based on prior knowledge, we seek to utilize the comparative and integrative capabilities of LLMs to inherently extract the fine-grained common part and integrate consistency knowledge from different responses. As shown in Figure~\ref{fig:method_workflow}, for the multiple input responses, we concatenate them all and construct a prompt with an instruction asking the LLM to extract the major consensus of generated responses, then integrate and generate the final response. In this way, FSC measures commonality at a finer granularity, utilizing, and integrating the consistency knowledge from different responses, alleviating the coarse-grained limitations of the selection based self-consistency.

\subsection{Candidates filtering}
As shown in Figure~\ref{fig:method_workflow}, considering the varying quality of responses generated by LLM, we propose candidates filtering strategy to measure the consistency of generated responses at the sample level, eliminating responses with low consistency to ensure the overall quality of the candidate responses. Specifically, given the similarity function $\mathit{Sim}$, we can define a sample-level self-consistency score $\mathit{SC_{Sim}(i)}$ for each response $\mathit{i}$, given by $SC_{Sim}(i)=\frac{1}{N-1}\sum_{j=1,j \neq i}^{N}Sim(j, i)$. Here, $N$ represents the number
of responses, and we use ROUGE \citep{lin-2004-rouge} for our similarity function $\mathit{Sim}$. In the end, we take the Top-$k$ responses according to $\mathit{SC_{Sim}}$ as the filtered candidates set, where $k$ is a hyperparameter.

\subsection{Merge}
As shown in Figure~\ref{fig:method_workflow}, to reduce the computation cost of FSC, we propose merging semantically similar candidates to decrease the number of responses to be integrated by the LLM. Furthermore, due to the limitations of the LLM input length, the number of samples inputted to the LLM for consistency extraction is limited. By merging, we can provide the LLM with samples that contain more diverse knowledge, broadening the LLM's knowledge field of view. specifically, To merge similar responses, we employ the K-Medoids clustering algorithm based on their semantic similarity. We categorize all responses into $c$ clusters, each encompassing a set of similar results. Then we select the centroids of each cluster as representative responses and discard
the remaining ones. It ensures the selected response has diverse knowledge and reduces the cost of FSC. 

\section{Experiment}

\subsection{Evaluation setup}
\subsubsection{Benchmarks}
We evaluate FSC on the following variety of tasks:

\paragraph{Code generation} We conduct experiments on three widely used code generation benchmarks, including HumanEval \citep{humaneval}, HumanEval+ \citep{humanevalplus} for Python code generation and BIRD-SQL dataset \citep{birdsql} for text-to-SQL generation. HumanEval \citep{humaneval} is a hand-written Python programming problem, which is further enhanced by HumanEval+ \citep{humanevalplus} through the addition of more unit tests. BIRD-SQL \citep{birdsql} is a much more challenging dataset consisting of text-to-SQL tasks across 37 professional domains, derived from 95 databases with a total size of 33.4 GB. 

\paragraph{Text summarization}
We conduct our experiments on two widely used text summarization datasets: DailyMail for short text summarization \citep{nallapati-etal-2016-abstractive} and SummScreen for long text summarization \citep{chen-etal-2022-summscreen}. In the DailyMail dataset, each input is an $\sim800$ words news article, and each reference output is a human-written summary of the article with $\sim55$ words. In SummScreen, every input is a transcript of a TV show episode with $\sim5,600$ words, and each reference output is a $\sim100$ words human-written recap of the episode. We follow \citet{nallapati-etal-2016-abstractive} and measure ROUGE 1, ROUGE 2, and ROUGE-L\footnote{ We use the implementation of \url{https://github.com/pltrdy/rouge.}} which measure n-gram overlap with the reference summary, and we also measure BERTScore F1\footnote{We use the bert-base-uncased version for evaluation.} \citep{bertscore}.

\definecolor{lightgreen}{RGB}{48,128, 20} 
\definecolor{lightred}{RGB}{220, 54, 20} 

\begin{table*}[ht]
\centering
\small
\begin{tabular}{clcccc}
\toprule
\multicolumn{1}{c}{\multirow{2}{*}{Model}} & \multicolumn{1}{c}{\multirow{2}{*}{Method}} & HumanEval & HumanEval+ & \multicolumn{2}{c}{BIRD-SQL} \\ \cmidrule{3-6} 
\multicolumn{1}{c}{} & \multicolumn{1}{c}{} & Accuracy & Accuracy & Accuracy & Efficiency \\ \toprule
\multirow{4}{*}{GPT-3.5-turbo} & Random & 69.8 \hspace{0.23cm}-\hspace{0.23cm} & 63.2 \hspace{0.23cm}-\hspace{0.23cm} & 43.1 \hspace{0.23cm}-\hspace{0.23cm} & 43.9 \hspace{0.23cm}-\hspace{0.23cm} \\
 & UCS & 70.1 \textcolor{lightgreen}{$\uparrow$0.3} & 63.2 \textcolor{lightgreen}{$\uparrow$0.0} & 43.6 \textcolor{lightgreen}{$\uparrow$0.5} & 44.5 \textcolor{lightgreen}{$\uparrow$0.6} \\
 & USC & 70.5 \textcolor{lightgreen}{$\uparrow$0.4} & 66.3 \textcolor{lightgreen}{$\uparrow$3.1} & 43.6 \textcolor{lightgreen}{$\uparrow$0.5}  & 44.6 \textcolor{lightgreen}{$\uparrow$0.7} \\
 & FSC & \textbf{74.5} \textcolor{lightgreen}{$\uparrow$4.7} & \textbf{68.4} \textcolor{lightgreen}{$\uparrow$5.2} & \textbf{45.3} \textcolor{lightgreen}{$\uparrow$2.2} & \textbf{46.0} \textcolor{lightgreen}{$\uparrow$2.1} \\ \midrule
\multirow{4}{*}{GPT-4} & Random & 82.5 \hspace{0.23cm}-\hspace{0.23cm} & 76.4 \hspace{0.23cm}-\hspace{0.23cm} & 49.5 \hspace{0.23cm}-\hspace{0.23cm} & 50.6 \hspace{0.23cm}-\hspace{0.23cm} \\
 & UCS & 82.3 \textcolor{lightred}{$\downarrow$0.2}  & 76.7 \textcolor{lightgreen}{$\uparrow$0.3} & 50.5 \textcolor{lightgreen}{$\uparrow$1.0} & 51.8 \textcolor{lightgreen}{$\uparrow$1.2} \\
 & USC & 86.1 \textcolor{lightgreen}{$\uparrow$3.6} & 80.9 \textcolor{lightgreen}{$\uparrow$4.5} & 50.9 \textcolor{lightgreen}{$\uparrow$1.4}& 51.6 \textcolor{lightgreen}{$\uparrow$1.0} \\
 & FSC & \textbf{87.1} \textcolor{lightgreen}{$\uparrow$4.6} & \textbf{82.8} \textcolor{lightgreen}{$\uparrow$6.4} & \textbf{51.4} \textcolor{lightgreen}{$\uparrow$1.9} & \textbf{52.2} \textcolor{lightgreen}{$\uparrow$1.6} \\ \bottomrule
\end{tabular}
\caption{The results on three code generation benchmarks. The improvements are calculated between each methods and Random. The best performance for each dataset are shown in \textbf{bold}.}
\label{tab: code}
\end{table*}

\paragraph{Mathematical reasoning} 
We introduce the widely used GSM8K \citep{GSM8K} and MATH \citep{MATH} datasets to verify the generalizability of the proposed method on tasks with answer of fixed form. GSM8K consists of 8,500 grade school math word problems, and MATH consists of 12,500 challenging mathematics problems from high school competitions.

\subsubsection{Baselines}
We compare FSC to the following self-consistency methods:
(1) Random selects one answer randomly from multiple responses with temperature > 0; (2) UCS \citep{UCS}\footnote{As token probabilities cannot be obtained from GPT-3.5-turbo and GPT-4, we implemented UCS based on its unigram version.}  calculates the overlap of unigrams between candidates and then selects the final answer with highest value; (3) USC \citep{USC} utilizes LLMs to choose the most consistent one as the final answer. (4) SC \citep{SC} is the standard self-consistency decoding with answer extraction, which mitigates noise from individual sampling by employing a voting mechanism across multiple samples. Specifically, random select represents the performance of the LLM itself when the temperature > 0, while UCS and USC are two strong baselines for selection based self-consistency methods. Since the outputs of code generation and summarization tasks are in free-form, we evaluate SC on mathematical reasoning benchmarks where the final answers can be compared through exact match.

\begin{table*}[ht]
\centering
\small
\begin{tabular}{llcccccccc}
\toprule
\multirow{2}{*}{Model} & \multirow{2}{*}{Method} & \multicolumn{4}{c}{DailyMail} & \multicolumn{4}{c}{Summscreen} \\ \cmidrule{3-10} 
 &  & Rouge1 & Rouge2 & RougeL & BertScore & \multicolumn{1}{c}{Rouge1} & \multicolumn{1}{c}{Rouge2} & \multicolumn{1}{c}{RougeL} & \multicolumn{1}{c}{BertScore} \\ \toprule
\multirow{4}{*}{GPT-3.5-turbo} & Random & 37.3 & 14.1 & 38.4 & 60.6 & 18.3 & 2.1 & 16.8 & 48.8 \\
 & UCS & 36.9 & 14.0 & 38.2 & 60.5 & 16.9 & 2.0 & 16.4 & 48.3 \\
 & USC & 37.7 & 14.3 & 38.7 & 60.8 & 19.3 & \textbf{2.3} & 17.2 & 49.3 \\
 & FSC & \textbf{38.6} & \textbf{14.4} & \textbf{39.0} & \textbf{60.9} & \textbf{20.1} & 2.1 & \textbf{17.4} & \textbf{49.6} \\ \midrule
\multirow{4}{*}{GPT-4} & Random & 37.5 & 14.5 & 38.9 & 61.0 &  19.1 & 2.6 & 17.3 & 49.5 \\
 & UCS & 36.9 & 14.3 & 38.6 & 60.9 & 18.6 & 2.4 & 17.0 & 49.3 \\
 & USC & 37.6 & 14.6 & 39.0 & 61.1 & 19.3 & 2.7 & 17.5 & 49.6 \\
 & FSC & \textbf{38.1} & \textbf{14.7} & \textbf{39.3} & \textbf{61.2} & \textbf{19.5} & \textbf{2.9} & \textbf{17.8} & \textbf{50.0} \\ \bottomrule
\end{tabular}

\caption{Results on summarization benchmarks. FSC consistently improves over the baselines on summary quality.}
\label{tab: summarization}
\end{table*}

\begin{table}[ht]
\centering
\small
\begin{tabular}{lccc}
\toprule
Dataset & UCS & USC & FSC \\ \toprule
DailyMail & 19.5\% & 26.5\% & \textbf{54.0\%} \\
Summscreen & 19.3\% & 31.5\% & \textbf{49.2\%} \\ \bottomrule
\end{tabular}
\caption{Comparison of GPT-4 score between different methods on GPT-3.5-turbo. For each test data, we use GPT-4 to score the quality of summaries generated by each method and count the proportion of the highest evaluation values obtained by each method on the entire dataset.}
\label{tab: gpteval}
\end{table}

\subsubsection{Implementation details}
We conduct experiments using GPT-3.5-turbo and GPT-4 models\footnote{we use the "2023-05-15" version of API for both.}. We set the temperature as 0.8 for both GPT-3.5-turbo (ChatGPT)\footnote{\url{https://chat.openai.com}} and GPT-4 \citep{gpt4} models to generate 50 initial responses for all benchmarks. For summarization and python code generation, the initial samples are generated with zero-shot prompting, thus the output formats are diverse. For BIRD-SQL, we used the 1-shot chain-of-thought prompt following \citet{birdsql}, which improves the performance. Considering the cost of experiments, we randomly select 1,000 samples from the test splits of DailyMail and SummScreen respectively to form our text summarization benchmarks. We set the temperature of FSC and USC as 0 to ensure the reproducibility of the results. Unless otherwise specified, we set the default number of input responses as 5 for all baselines. All experiments are repeated five times and the average performance is reported.

\subsection{Main results}

\begin{table}[ht]
\centering
\small
\begin{tabular}{llcc}
\toprule
Model & Method & \multicolumn{1}{l}{GSM8K} & \multicolumn{1}{l}{MATH} \\ \toprule
\multirow{5}{*}{GPT-3.5-turbo} & Random & 75.9 & 35.0 \\
 & UCS & 77.2 & 35.6 \\
 & USC & 80.6 & 39.3 \\
 & SC & \textbf{82.0} & \textbf{41.9} \\
 & FSC & \underline{81.0} & \underline{39.5} \\ \midrule
\multirow{5}{*}{GPT-4} & Random & 87.5 & 50.2 \\
 & UCS & 87.6 & 50.7 \\
 & USC & 88.1 & 54.8 \\
 & SC & \underline{88.8} & \underline{55.5} \\
 & FSC & \textbf{91.3} & \textbf{56.0} \\ \bottomrule
\end{tabular}
\caption{Accuracy on mathematical reasoning benchmarks. FSC performance is comparable to SC. The best and second best performance for each dataset are shown in \textbf{bold} and \underline{underline}. }
\label{tab: mathematical}
\end{table}

\begin{table*}[ht]
\centering
\small
\begin{tabular}{lccccccc}
\toprule
\multirow{2}{*}{Method} & \multicolumn{1}{l}{\multirow{2}{*}{N}} & HumanEval & HumanEval+ & \multicolumn{4}{c}{DailyMail} \\ \cmidrule{3-8} 
 & \multicolumn{1}{l}{} &  Accuracy & Accuracy & Rouge1 & Rouge2 & RougeL & BertScore \\ \toprule
UCS & 10 & 70.7 & 62.5 & 38.2 & 14.4 & 39.2 & 60.9 \\
USC & 10 & 71.0 & 65.9 & 38.5 & 14.5 & 39.1 & \textbf{61.1} \\
FSC & 5 & 74.5 & 68.4 & 38.6 & 14.4 & 39.0 & 60.9 \\
FSC+Filter & 10 & \textbf{75.5} & \textbf{69.2} & \textbf{38.7} & \textbf{14.6} & \textbf{39.3} & 61.0 \\ \bottomrule
\end{tabular}
\caption{Results of candidates filtering on HumanEval, HumanEval+ and DailyMail with GPT-3.5-turbo. N represents the number of input responses. The best performance for each dataset is shown in \textbf{bold}.}
\label{tab: ablation_filter}
\end{table*}

\begin{table*}[ht]
\setlength{\tabcolsep}{0.54em} 
\centering
\small
\begin{tabular}{llcccccccc}
\toprule
\multirow{2}{*}{Method} & \multirow{2}{*}{N} & HumanEval & HumanEval+ & \multirow{2}{*}{Save} & \multicolumn{4}{c}{DailyMail} & \multirow{2}{*}{Save} \\ \cmidrule{3-4} \cmidrule{6-9}
 &  & Accuracy & Accuracy &  & Rouge1 & Rouge2 & RougeL & BertScore &  \\ \toprule
FSC & 5 & 74.5 & 68.4 & - & 38.6 & \textbf{14.4} & \textbf{39.0} & \textbf{60.9} & - \\
FSC & 4 & 74.3 & 67.8 & 20.0\% & \textbf{38.8} & 14.2 & \textbf{38.8} & 60.8 & 20.0\% \\
FSC+Merge & 5 & 75.3 & 68.0 & 25.4\% & \textbf{38.8} & 14.0 & 38.6 & 60.8 & 26.5\% \\
FSC+Filter+Merge & 5 & \textbf{75.6} & \textbf{68.5} & \textbf{40.0\%} & 38.6 & \textbf{14.4} & 38.9 & \textbf{60.9} & \textbf{53.2\%} \\ \bottomrule
\end{tabular}
\caption{Results of merge on HumanEval, HumanEval+ and DailyMail with GPT-3.5-turbo. "Save" represents how many candidate responses input into the LLM are saved compared to the default settings of FSC (N=5). We calculate "Save" through relative difference percentage. The best performance for each dataset is shown in \textbf{bold}.}
\label{tab: ablation_merge}
\end{table*}

\subsubsection{Code generation}
As shown in Table \ref{tab: code}, we present the results on HumanEval, HumanEval+ and BIRD-SQL respectively. Besides the execution accuracy, we follow \citet{birdsql} to also evaluate the valid efficiency score on BIRD-SQL, which measures the efficiency of the generated SQL queries. We show that FSC outperforms all baselines on execution accuracy by a significant margin across all datasets, while also generating more efficient SQL code.

\subsubsection{Text summarization}
Table \ref{tab: summarization} presents the results for summarization benchmarks. In both datasets FSC consistently improves over the baselines across all metrics, which demonstrates that FSC can improve performance in both short and long text summarization tasks simultaneously. Considering that LLMs demonstrate better consistency with humans in evaluation tasks \citep{liu-etal-2023-g, yuan2023batcheval}, we employ GPT-4 as an evaluator to assess the generated summaries, following  \citet{liu-etal-2023-g}. As shown in Table \ref{tab: gpteval}, the experimental results indicate that FSC is superior to both UCS and USC.

\subsubsection{Mathematical reasoning}
As shown in Table \ref{tab: mathematical}, besides selection based self-consistency methods, we compare FSC against the standard self-consistency (SC). For SC, we employ a regular expression matching to extract the final answer on GSM8K, and reuse the answer parsing code from \citet{li2023escape} for MATH. Overall, FSC consistently improves over the selection based self-consistency methods UCS and USC, and the performance is generally comparable to SC, which needs answer parsing to perform the voting. surprisingly, FSC outperform SC on GPT-4, which demonstrates that FSC is not simply dependent on statistical measures of final reasoning answers, and its analysis and integration of various reasoning paths are effective. These results suggest that FSC could be further generalized to tasks where answer extraction is feasible for voting.


\begin{figure*}[ht]
\begin{center}
\small
\includegraphics[width=0.88\textwidth]{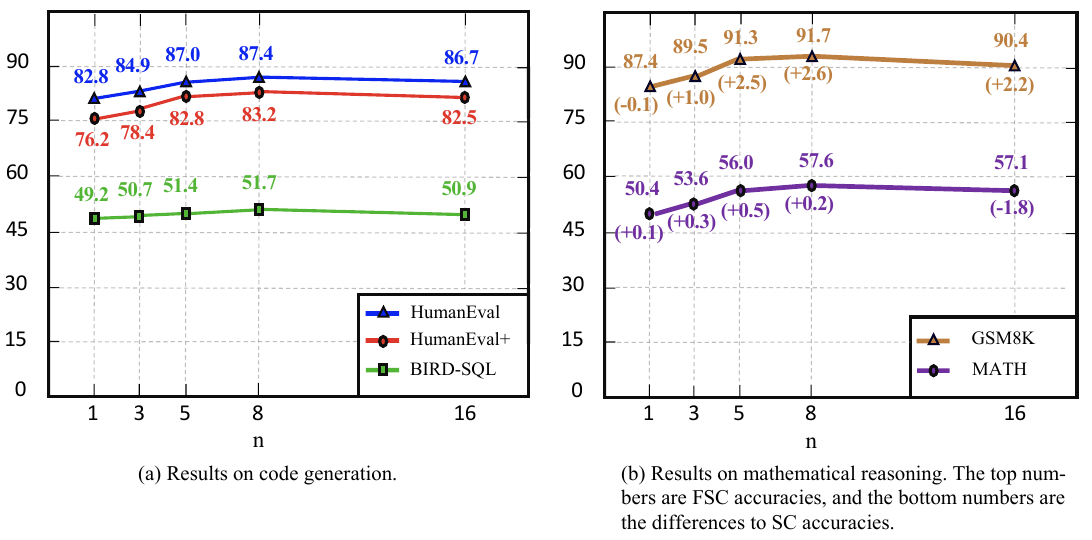}
\end{center}
\caption{FSC results on GPT-4 model with different number of input responses.}
\label{fig:experiment_fusion_num}
\end{figure*}

\subsection{Effect of candidates filtering}
As shown in Table \ref{tab: ablation_filter}, we compare the performance of FSC combined with candidates filtering strategy (denoted as FSC+Filter) with FSC itself and selection based self-consistency baselines. Specifically, FSC+Filter performs candidate filtering on the initial $N$ responses to obtain $\frac{N}{2}$ filtered responses, and then applies FSC to the filtered responses to get the final output. The results show that candidates filtering consistently improves FSC performance across all test benchmarks, indicates that candidates filtering obtains a higher-quality response candidate set through screening. On the other hand, the performance of FSC+Filter surpasses UCS and USC under the same number of response inputs on all test datasets (except for BertScore on DailyMail), demonstrating the superiority of FSC combined with candidates filtering strategy.

\subsection{Effect of merge}
We present the results of FSC combined with merge strategy in Table \ref{tab: ablation_merge}. The results show that the Merge strategy can significantly reduce the number of FSC input responses (20.0\% on HumanEval and HumanEval+; 26.5\% on Daily Mail), with minimal performance loss. Compared to saving costs by reducing the number of input responses ($N=4$ for FSC), the Merge strategy saves more costs and maintains better performance (except for Rouge2 and RougeL on DailyMail), demonstrating its effectiveness. Furthermore, we combine FSC with both Filter and Merge strategies and achieve the best performance on HumanEval and HumanEval+, saving 40.0\% of the cost. Besides, we save 53.2\% of the cost on DailyMail with minimal performance drop. The results demonstrate the superiority of the combination of two proposed strategies.

\begin{table*}[ht]
\centering
\small
\begin{tabular}{ccccccccc}
\toprule
\multirow{2}{*}{Distribution} & \multicolumn{4}{c}{GPT-3.5-turbo} & \multicolumn{4}{c}{GPT-4} \\ \cmidrule{2-9} 
 & Random & UCS & USC & FSC & Random & UCS & USC & FSC \\ \toprule
5 / 0 & 0.0\% & 0.0\% & 0.0\% & \textbf{\textcolor{lightgreen}{1.1\%}} & 0.0\% & 0.0\% & 0.0\% & \textbf{\textcolor{lightgreen}{2.2\%}} \\
4 / 1 & 20.0\% & \textcolor{lightred}{13.9\%} & \textcolor{lightred}{13.9\%} & \textbf{\textcolor{lightgreen}{27.9\%}} & 20.0\% & \textcolor{lightgreen}{23.1\%} & \textcolor{lightred}{16.6\%} & \textbf{\textcolor{lightgreen}{29.1\%}} \\
3 / 2 & 40.0\% & \textcolor{lightred}{35.8\%} & \textcolor{lightgreen}{47.1\%} & \textbf{\textcolor{lightgreen}{62.2\%}} & 40.0\% & \textcolor{lightred}{31.0\%} & \textcolor{lightgreen}{51.7\%} & \textbf{\textcolor{lightgreen}{55.1\%}} \\
2 / 3 & 60.0\% & \textcolor{lightgreen}{68.3\%} & \textcolor{lightred}{56.6\%} & \textbf{\textcolor{lightgreen}{70.0\%}} & 60.0\% & \textcolor{lightgreen}{65.9\%} & \textbf{\textcolor{lightgreen}{87.2\%}} & \textcolor{lightgreen}{82.9\%} \\
1 / 4 & 80.0\% & \textcolor{lightgreen}{82.1\%} & \textcolor{lightgreen}{86.9\%} & \textbf{\textcolor{lightgreen}{91.6\%}} & 80.0\% & \textcolor{lightred}{73.6\%} & \textcolor{lightgreen}{91.6\%} & \textbf{\textcolor{lightgreen}{97.2\%}} \\ \bottomrule
\end{tabular}
\caption{Accuracy on code generation benchmark HumanEval with GPT-3.5-turbo and GPT-4.  Please refer to Table \ref{tb:intro} for the definition of distribution. We mark values lower than random performance in \textcolor{lightred}{red} and values higher than random in \textcolor{lightgreen}{green}. The best performance is highlighted in \textbf{bold}.}
\label{tab: distribution analysis}
\end{table*}

\subsection{Discussion}
\subsubsection{Different number of input responses} \label{sec: number of responses}
As shown in Figure \ref{fig:experiment_fusion_num}, we examine the effect of using different numbers of responses (denoted as $n$) in FSC on GPT-4 model.\footnote{Due to budget constraints, we do not conduct the experiment on text summarization benchmark.} The results show that FSC consistently benefits from more input responses with $n\leq8$. However, the performance of FSC decreases with $n=16$, which could be due to the difficulty in understanding long-context when the prompt includes a larger number of input responses, as well as the length constraint of LLMs. Nevertheless, we believe that using a limited number of input responses (e.g., 5) strikes an ideal balance between task accuracy and computational cost. In such case, FSC reliably enhances performance across all benchmarks. 

\begin{table}[t]
\centering
\small
\begin{tabular}{lcc}
\toprule
Method & \multicolumn{1}{l}{HumanEval} & \multicolumn{1}{l}{HumanEval+} \\ \toprule
Random & 19.7 & 16.9 \\
UCS & 20.3 & 17.2 \\
USC & 26.2 & 21.9 \\
FSC & \textbf{29.9} & \textbf{24.4} \\ \bottomrule
\end{tabular}
\caption{Accuracy on HumanEval and HumanEval+ with Mistral-7B-Instruct-v0.2.}
\label{tab: open-source}
\end{table}

\subsubsection{Robustness to noise in input responses}
Table \ref{tab: distribution analysis} shows the accuracy on the HumanEval benchmark under different distributions. We define noise as the the proportion of erroneous examples in the input responses, (Distribution 5/0 corresponds to the maximum noise). While the performance of UCS and USC lag behind random select when the input noise is high, FSC consistently surpasses random select by a large margin under all distributions, proving that FSC has better robustness. Furthermore, the accuracy of FSC is greater than 0 when the distribution is 5/0, indicating that FSC can still recover the correct answer when all input responses are wrong. This demonstrates that FSC is capable of integrating the correct knowledge from different input responses and eliminating the wrong part to achieve the correct final response.

\subsubsection{Generalizability on open-source small model}
As shown in Table \ref{tab: open-source}, We validate the generalization of FSC on the open-source small model Mistral-7B-Instruct-v0.2\footnote{\url{https://huggingface.co/mistralai/Mistral-7B-Instruct-v0.2}} in code generation tasks. We set the temperature for baseline sampling to 0.2 and kept the rest of the implementation completely consistent with the main experiment. The experimental results indicate that FSC has the potential to work effectively on smaller models.

\subsubsection{Segment-level consensus of FSC}
To provide additional evidence that FSC can incorporate a higher level of segment-level consensus, we carry out quantitative experiments. For all the generated candidates, we construct a pool of 4-grams (as representations of segments), and then calculate the overlap between the 4-grams of the final sample and the pool. We compare our method against two key baselines by computing the win rate. As shown in Table \ref{tab: fine-grained consensus}, the results demonstrate that our method can integrate more fine-grained segments from the candidate set, thereby generating samples of higher quality.

\begin{table}[t]
\centering
\small
\begin{tabular}{lcc}
\toprule
Win rate & FSC vs UCS & FSC vs USC \\ \toprule
DailyMail & 79.2\% & 67.1\% \\
Summscreen & 87.6\% & 81.8\% \\ \bottomrule
\end{tabular}
\caption{Comparison of fine-grained consensus obtained by different methods with GPT-4.}
\label{tab: fine-grained consensus}
\end{table}
\section{Conclusion}
In this work, we propose Fine-Grained Self-Consistency (FSC), which fully leverages the segment-level consensus  knowledge extracted within multiple samples to overcome the coarse-grained limitations faced by selection based self-consistency methods. To improve performance and minimize costs, we further propose two strategies called candidates filtering and merge. Extensive experiments demonstrate that FSC notably boosts the performance on diverse range of tasks, exhibits superior robustness against noise in input responses, and can be generalized to those tasks where answer extraction is feasible through voting. Additional experiments confirm that the proposed candidates filtering and merge strategies can further enhance the performance of FSC while reducing the required computational cost.

\section*{Limitations}
Despite the remarkable performance gain on variety of tasks, the current implementation of FSC still suffers from the following limitation:
As illustrated in Section \ref{sec: number of responses}, While self-consistency can be applied to any number of samples, the number of samples supported by FSC is bounded by the context length of the underlying LLM. That is to say, FSC would be limited in tasks that require lengthy responses, such as story generation, long text translation, etc. 
 
\section*{Ethics Statement}
All of the datasets used in this study were publicly available, and no annotators were employed for our data collection. We confirm that the datasets we used did not contain any harmful content and was consistent with their intended use (research). We have cited the datasets and relevant works used in this study.
\section*{Acknowledgements}
This work is supported by the Beijing Natural Science Foundation, China (Nos. 4222037, L181010). We thank the anonymous reviewers for their constructive comments.


\bibliography{anthology,custom}
\clearpage

\appendix
\section{Appendix}

\subsection{Performance of FSC under different temperatures}
To verify the generalizability of FSC under different temperature settings, we conduct experiments based on GPT-3.5-turbo under different temperature settings on HumanEval+ dataset. As shown in Table \ref{tab: different temperatures}, the results show that FSC exhibits consistent generalization under different temperature settings, with more significant performance improvements compared to the baselines when the temperature is higher. We hypothesize that this could be due to the fact that as the temperature increases, the diversity of the samples also increases, thereby enriching the knowledge from the input samples that FSC is able to integrate.

\begin{table}[h]
\centering
\small
\begin{tabular}{lccccc}
\toprule
Temperature & 0.2 & 0.4 & 0.6 & 0.8 & 1.0 \\ \toprule
Random & 63.5 & 64.1 & 64.2 & 63.2 & 63.6 \\
UCS & 63.2 & 63.9 & 64.1 & 63.2 & 64.2 \\
USC & 64.9 & 65.7 & 66.9 & 66.5 & 66.8 \\
FSC & \textbf{66.1} & \textbf{66.9} & \textbf{68.4} & \textbf{68.4} & \textbf{69.6} \\ \bottomrule
\end{tabular}
\caption{Accuracy on HumanEval+ under different temperatures with GPT-3.5-turbo.}
\label{tab: different temperatures}
\end{table}


\subsection{Sampling cost of FSC}
Under the setting of input sample size N=10, we conduct a statistical analysis of the token cost\footnote{we convert the token cost into price according to \url{https://openai.com/pricing} and report the average cost for every thousand samples.} on the HumanEval+ dataset based on GPT-3.5-turbo. As shown in Table \ref{tab: sampling cost}, the results show that the token cost of FSC+filter+merge is comparable to that of USC, while FSC achieves a significant performance improvement.

\begin{table}[h]
\setlength{\tabcolsep}{3pt}
\centering
\small
\begin{tabular}{lcccc}
\toprule
Method & Prompt & Completion & Price & Acc \\ \toprule
UCS & - & - & 0 & 62.5 \\
USC & 1207 & 9 & 0.53 & 65.9 \\
FSC+filter+merge & 724 & 128 & 0.55 & 69.2 \\ \bottomrule
\end{tabular}
\caption{Comparison of the cost of USC, UCS and FSC+filter+merge.}
\label{tab: sampling cost}
\end{table}

\begin{table*}[ht]
\centering
\small
\begin{tabular}{lcccc}
\toprule
Model & Method & HumanEval & HumanEval+ & BIRD-SQL \\ \toprule
\multirow{2}{*}{GPT-3.5-turbo} & MBDR & 71.5 & 65.6 & 44.1 \\
 & FSC & 74.5 & 68.4 & 45.3 \\ \midrule
\multirow{2}{*}{GPT-4} & MBDR & 84.4 & 78.6 & 50.8 \\
 & FSC & 87.1 & 82.8 & 51.4 \\ \bottomrule
\end{tabular}
\caption{Comparison of accuracy between FSC and MBDR on code generation tasks.}
\label{tab: MBDR}
\end{table*}

\subsection{Case study of FSC}
\label{ap: final response}
\begin{figure*}[ht]
\begin{center}
\includegraphics[width=1.0\textwidth]{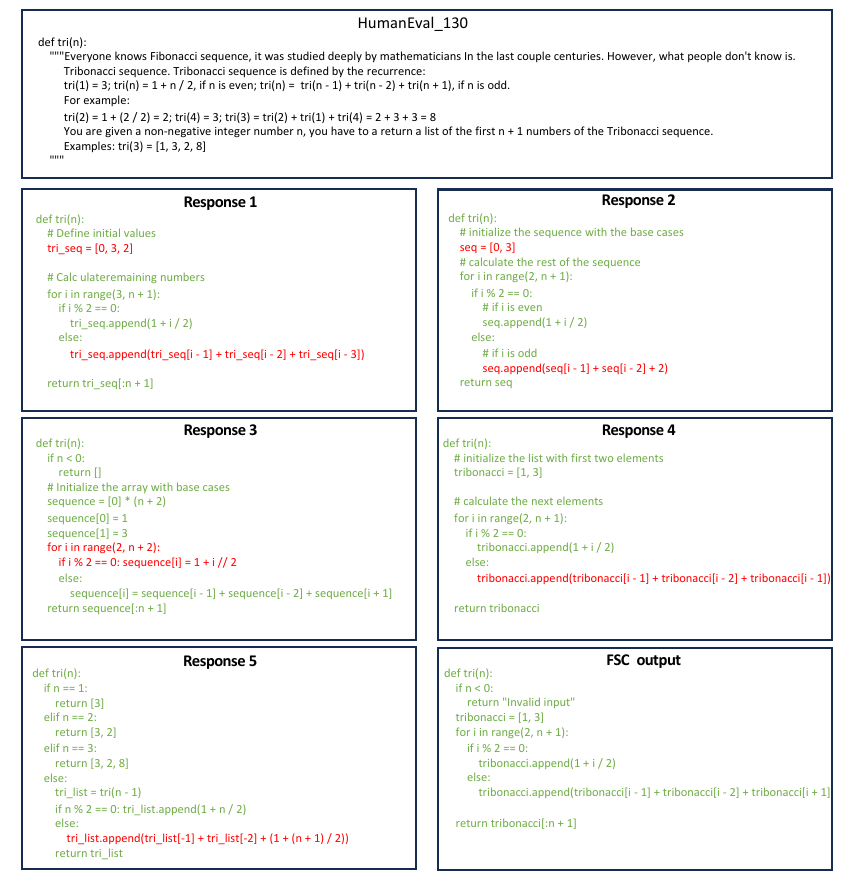}
\end{center}
\caption{The input responses and final output of FSC with GPT-3.5-turbo on HumanEval\_130. We mark the correct parts in \textcolor{lightgreen}{green} and the wrong parts in \textcolor{lightred}{red} for each input response.}
\label{fig:appendix_case_code}
\end{figure*}

To gain a more intuitive understanding of the working mechanism of FSC, and to conduct a qualitative analysis of the consistency of FSC's final output, we present the case of FSC on HumanEval benchmark. Figure \ref{fig:appendix_case_code} shows the final output of FSC on HumanEval\_130 when the input responses are all wrong. Specifically, for all error input responses, FSC incorporates consensus knowledge from each input and eliminates low-quality segments, ultimately recovering the correct solution. Our analysis indicates that FSC is capable of achieving fine-grained commonality extraction and obtaining outputs with better consistency compared to selection based self-consistency methods.

\subsection{Comparision with Minimum Bayes Risk Decoding (MBDR)}
\citet{suzgun-etal-2023-follow} propose MBDR method, achieving sample selection by calculating the BertScore between the generated samples. As shown in Table \ref{tab: MBDR}, we reproduce the MBDR according to the original paper, and compare it with FSC on our code generation benchmark.

\subsection{Input document for Figure~\ref{fig:intro_case}}
\label{ap:input}
Input: "The wedding of the year in Scotland takes place on Saturday when British No 1 and two-time Grand Slam champion Andy Murray marries Kim Sears, his girlfriend of almost 10 years, in his hometown of Dunblane. Murray and Sears, both aged 27, met when the pair were teenagers during the US Open in 2005. Murray was playing in only his second Grand Slam tournament, while Sears was travelling with her tennis coach father Nigel. The couple got back together in 2010 following a brief split and after having to field constant questions over the years on when he would propose, their engagement was confirmed last November. Andy Murray kisses his new girlfriend Kim Sears in the crowd after winning his first ATP World Tour title in San Jose in February 2006 . Murray pictured walking alongside Sears on the streets of London during the Wimbledon Championships in June 2006 . Murray watches his brother Jamie in action at London's Queen's Club in June 2007 alongside Sears and mother Judy (left) Murray watches his brother in action at Wimbledon in 2007 alongside Sears and Carlos Mier (right), who will be one of Murray's three best men . Murray and Sears watching British boxer Amir Khan in action during a title fight at the ExCel Arena in London in February 2008 . Murray and Sears attend the exhibition match held to mark the launch of the new Wimbledon Centre Court roof in May 2009 . Murray and Sears attend a Burberry fashion show alongside Serena Williams (second left) and Sarah Jessica Parker (left) in September 2010 . Murray was the best man for the wedding of his brother Jamie (right) and wife Alejandra (second right) at Cromlix House in October 2010 . Television viewers are well used to the sight of Sears, pictured here at Wimbledon in 2011, showing her emotions during Murray's matches . Murray looks dejected as he and Sears wait for transport after the Brit lost the Wimbledon 2012 final to Roger Federer . There were happier moments just weeks later though as Murray celebrates with Sears after beating Federer to win Olympic gold in London . Murray then won his first Grand Slam title at the US Open in New York in September 2012 by beating Novak Djokovic in an epic final . Murray and Sears laugh with television host Jimmy Fallon before an appearance on the show following his US Open victory . Murray and Sears pose for photographers as they arrive for a Burberry fashion show during London Fashion Week in September 2012 . Murray leans over to kiss Sears after becoming the first British man in 77 years to win the men's singles title at Wimbledon in July 2013 . Murray and Sears pose with the famous trophy during the Wimbledon Champion's Dinner at a hotel in Park Lane later that evening . Murray and Sears stand outside Buckingham Palace in October 2013 after the British No 1 was awarded an OBE by Prince William . Murray and Sears outside Dunblane High School after the local hero received the Freedom of Stirling at his former school in April 2014 . Murray and Sears watch golf as the couple stroll by the fairways of Ridgewood Country Club in New Jersey during The Barclays in August 2014 . Sears poked fun at the reaction to her foul-mouthed rant during the Australian Open this year by wearing this t-shirt for Murray's final."

\end{CJK}
\end{document}